# Detecting AI Generated Text Based on NLP and Machine Learning Approaches

*Nuzhat Noor Islam Prova*
*Department of Seidenberg School of CSIS*
*Pace University*
*NewYork, USA*
*nuzhatnsu@gmail.com*

*Abstract*— **Recent advances in natural language processing (NLP) may enable artificial intelligence (AI) models to generate writing that is identical to human written form in the future. This might have profound ethical, legal, and social repercussions. This study aims to address this problem by offering an accurate AI detector model that can differentiate between electronically produced text and human-written text. Our approach includes machine learning methods such as XGB Classifier, SVM, BERT architecture deep learning models. Furthermore, our results show that the BERT performs better than previous models in identifying information generated by AI from information provided by humans. Provide a comprehensive analysis of the current state of AI-generated text identification in our assessment of pertinent studies. Our testing yielded positive findings, showing that our strategy is successful, with the BERT emerging as the most probable answer. We analyze the research's societal implications, highlighting the possible advantages for various industries while addressing sustainability issues pertaining to morality and the environment. The XGB classifier and SVM give 0.84 and 0.81 accuracy in this article, respectively. The greatest accuracy in this research is provided by the BERT model, which provides 0.93% accuracy.**

## I. Introduction

In the rapidly evolving fields of artificial intelligence (AI) and natural language processing (NLP), computers are already able to produce writing that is completely unlike anything that is written by a human. Although there are many potentials uses for this scientific competence, it has also led to a number of constitutional, legal, and societal issues. An artificial intelligence detector model that is specifically designed to differentiate between human-written text and text that is created programmatically. A new era of human-like writing pattern emulation by robots has begun with the development of huge language models. This has led to serious moral conundrums and necessitated a fundamental change in the way we interpret and interact with text. Addressing the critical need for a method to distinguish between content created by AI and content created by humans. Motivated by the need to navigate the ethical ambiguities present in writing produced by artificial intelligence. The introduction delineates the trajectory of our investigation, delving into the motivations that underscore the significance of our findings. The prevalence of artificial intelligence (AI)-generated content in several industries gives rise to concerns over its accuracy, dependability, and potential for manipulation. As we embark on this journey, several important questions come up: What distinguishes information generated by algorithms from that generated by human cognitive processes? What cultural effects results from this inability to discriminate? This introduction explains the logic underlying our AI sensor model, setting the stage for our research of these problems. In addition to providing a technological solution, our research aims to initiate a wider conversation on the ethical implications of AI-generated writing, opening the door for the responsible and proper use of advanced computational language models in the era of digital communication. Our study is motivated by the urgent need to provide a concrete and useful solution to the moral, societal, and legal challenges posed by the increasing complexity of AI-generated text. The threat of misinformation, content manipulation, and eroding trust intensifies when AI models have text generation abilities comparable to those of humans. The lack of a thorough process for differentiating material created by AI from that produced by humans impedes responsibility and openness. By developing an AI-detecting framework that can differentiate between artificially generated and human-crafted literature. The detection technique not only resolves existing moral and legal issues, but it also lays the foundation for promoting reasonable and approachable behavior in the expanding field of AI-generated interactions. Recently, AI language models have shown remarkable ability to generate text that mimics human writing. These models use massive data and advanced algorithms to generate coherent and contextually appropriate writing across many themes and styles. This has led to many applications in content generation, virtual assistants, and automated customer support, but it has also raised concerns about the misuse of AI-generated text for malicious purposes like misinformation, public opinion manipulation, and fraud. Thus, there is a rising need to identify and limit AI-generated material, especially on online platforms where it might be difficult to discern between human and AI-generated writing. Language's intricacy and small differences between human and machine-generated text make spotting AI-generated material difficult. AI language models excel in mimicking human language's syntactic and semantic structures, but they frequently show evidence of their non-human origin. AI-generated writing may be incoherent, inconsistent, or implausible, deviating from human language conventions. AI-generated writing may also expose training data biases or preferences, making it harder to discern between human and AI-authored material. NLP and machine learning methods for AI-generated text detection have been developed to overcome these issues. These methods analyze text's linguistic, statistical, or behavioral aspects to find abnormalities or departures from human-generated material. Stylometric analysis, anomaly detection, and adversarial testing are common methods. Stylometric analysis examines vocabulary usage, sentence structure, and writing style. Anomaly detection identifies textual anomalies compared to a baseline of human-authored content.

## II. Literature Review

The machine-learning-based algorithm developed in [1] against the popular text generation technique known as Generative Pre-trained Transformer (GPT) to see how well it could distinguish between texts produced by AI and writings created by humans. Using a Neural Network with three hidden layers and Small BERT, we get a high accuracy performance. The degree of precision attained varies and relies on the loss function used to the classification detection. In order to counteract disinformation and explore possible future research subjects, the goal of this effort is to assist future studies in text bot identification. Evaluation of a statistical language model for text steganography based on artificial intelligence is the main objective of this project, as stated in [2]. The advantageous properties of a Markov chain model based on natural language processing are suggested in this study for an autonomously produced cover text. This paper presents a comparative analysis of the steganographic embedding rate, volume, and other properties of two RNN steganographic schemes: RNN-generated Lyrics and RNN-Stega. The acronym for recurrent neural networks is RNN. Included in the essay is a case study about information security and artificial intelligence. This case study investigates the background, applications, challenges, and ways in which artificial intelligence (AI) might help mitigate security threats and weaknesses. As stated in [3], identifying machine-generated text is an essential preventative precaution against the misuse of natural language generation models yet, there are several unresolved issues and significant technical challenges in this area. An extensive analysis of threat models presented by contemporary NLG systems and the most thorough investigation of methods available for machine-generated text detection to date. To underline how crucial it is that detection systems themselves demonstrate their reliability by being impartial, strong, and responsible. In [4], it is said that the advancement of natural language processing methods and sarcasm detection technologies would enable intelligent and economical collaboration with machine devices as well as advanced human-machine interactions. In this work, introduce stance-level sarcasm detection a novel job. Finding the author's hidden position and using that information to ascertain the text's ironic polarity are the goals of this effort. The term "stance-level sarcasm detection" is abbreviated as "SLSD." Afterward, we provide a complete framework consisting of an innovative stance-centered graph attention network and Bidirectional Encoder Representations from Transformers (BERT). It is evident from the experiment results that the SCGAT framework outperforms the state-of-the-art baselines by a wide margin. In the [5] By using Natural Language Processing (NLP) and Deep Learning, it is possible to identify, explore, and develop a global comprehension of the emotions that were expressed in the first months of the COVID-19 pandemic. Transfer Learning and Robustly Optimized BERT Pretraining Approach, an advanced deep learning approach, were used to collect and analyze approximately two million tweets. The collection of these tweets took place between February and June of 2020.The standard Emotion Dataset from CrowdFlower, which was sourced from Reddit, was used to facilitate transfer learning and compiling the Twitter dataset, a multi-class emotion classification system was constructed. Compared to the previously used AI-based emotion classification techniques, they were able to achieve an accuracy of 80.33% in tweet categorization and an average MCC score of 0.78. The novel use of the Roberta model during the epidemic is described in this article. This program offered insights on the evolving mental health of several global residents throughout time. In the [6] This article presents an approach based on Arabic text recognition for detecting misogynistic words in Arabic tweets. Tests using the Arabic Levantine Twitter Dataset for Misogynistic revealed that the proposed technique obtained 90.0% and 89.0% recognition accuracies for binary and multi-class tasks, respectively. The method that was recommended seems to be useful in providing sensible and useful answers for identifying sexism in Arabic on social media. They describe the basic process of chatbots in [7] and compare them according to the technology approach used as well as a few other significant criteria. To offer a relevant response, a chatbot must precisely analyze and understand user input. This will make it possible to have more productive conversations using natural language. These days, they are used in many important domains, including research, education, and healthcare. The relationship between people and technology may be more natural with chatbots. The evolution of synthetic text generation has raised a number of crucial issues that might have an impact on society and the internet. Programs powered by LLM are expected to replace a significant portion of the human labor [8]. A plethora of tasks in the domains of research, education, law, advertising, and creative writing for leisure are now being performed by computers. It is more difficult to spot cases of phishing, spamming, academic fraud, fake news, and reviews [9]. As a result, it will be very difficult in the future to identify a phrase that has been falsely constructed. In order to identify unique traits and patterns in text material that has been intentionally generated, to look into a number of approaches in this research project. The problem of text recognition posed by artificial intelligence, or more broadly, machine-generated textual material, has drawn a lot of interest. It started with the Turing test and was then used to the evaluation of chatbots [10]. In this paper, computerized procedures were the major focus instead of hybrid or human-cantered detection techniques. According to Crothers, Japkowicz, and Viktor [11], autonomous AI-generated text recognition systems may be broadly classified into two categories: feature-based and neural language models. Nonetheless, specific domains were the focus of other research, such as [12] scientific settings and fake news false reviews misinformation. In addition to, two methods for automatically identifying text generated by AI were proposed in this study text resemblance-based approaches and feature-based approaches using machine learning models.

## III. Proposed Methodology

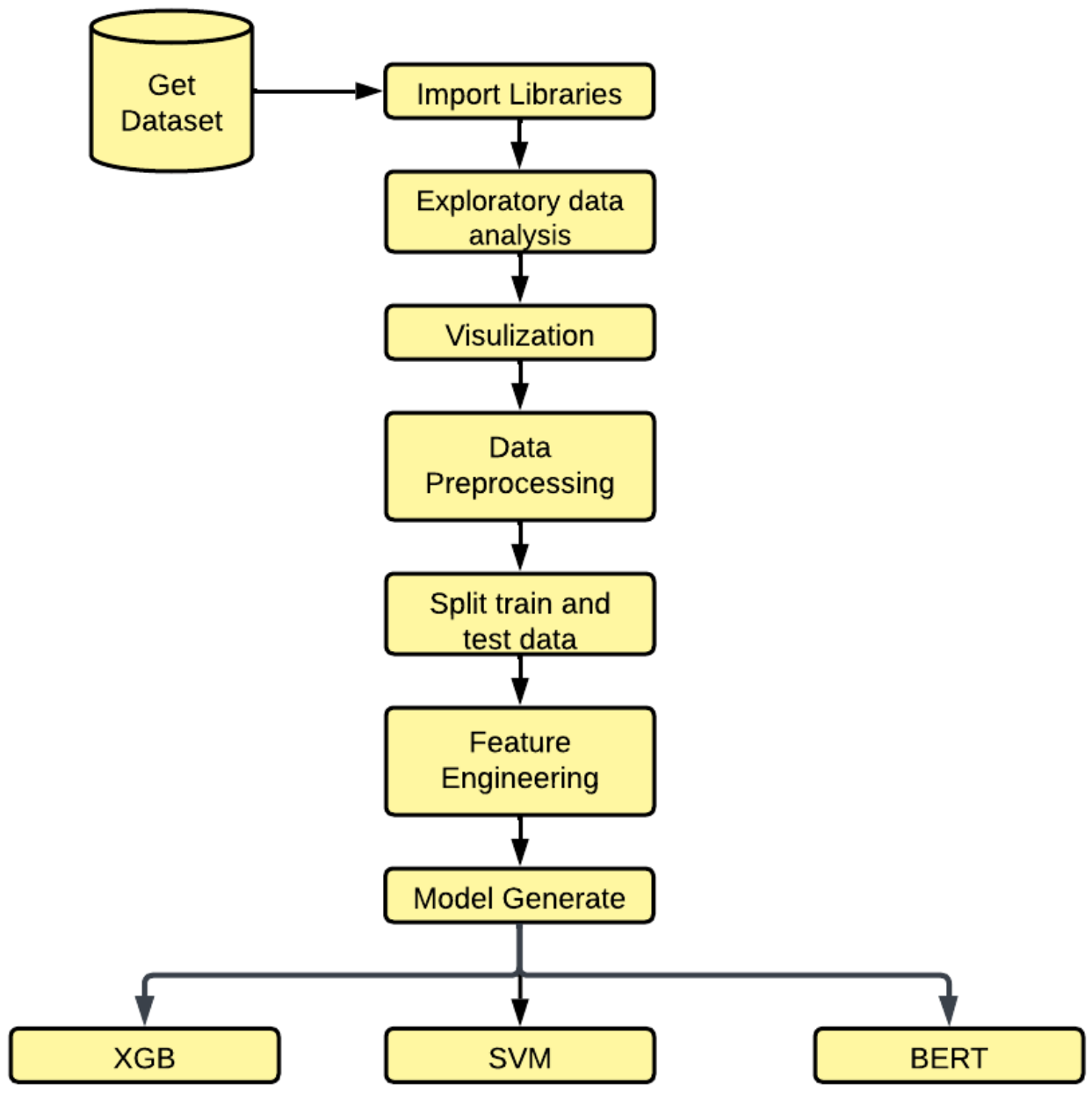

Fig. 1. The proposed methodology of the work

### A. Dataset

This dataset was essentially built by myself. which I separated into 0 and 1 levels. in which AI-generated data is ranked as 1, and human-generated data as 0. I have produced 3000 total data points, 1500 of which are produced by humans and 1500 by artificial intelligence. I used both handwritten and online platform. The datasets include text written by both AI and humans in order to provide a thorough understanding of distinguishing factors.

### B. Data Preprocessing

*Stopwords removal:*
Stop words are typical natural language terms that are filtered before or after text processing. In search engines, text mining, and machine learning, these phrases are usually redundant or non-informative. Stop words in English include "the", "is", "at", "which", "on", "in", "to", "of", "and", "or". These words are commonly deleted from text data to save computational overhead and concentrate on more significant terms to increase algorithm efficiency and accuracy.

*Remove unwanted characters or links:*
Preprocessing text data by removing unwanted letters and links is necessary for natural language processing, web scraping, and data analysis. Remove undesired special characters, punctuation, HTML components, and URLs. Regular expressions target and eliminate URLs and punctuation. For filtering, tokenization may split text into words or tokens. Filters eliminate unwanted text and leave just what's needed. Change all content to lowercase or expand contractions to standardize format. Quality assurance tests during cleaning save important information and eliminate useless items. Analysts and developers may ensure downstream processes and analytics are correct by cleaning and sanitizing text data.

*Stemming:*
Words are reduced to their stem, or basis, by the application of the stemming approach in natural language processing. By considering word variants as the same, stemming aims to normalize words with similar meanings to a common form, which may enhance text analysis and retrieval tasks. For instance, the basic form "run" would be stemmed to produce the words "running," "runs," and "ran."

*Lemmatization:*
Lemmatization is a term used in natural language processing to describe the process of learning words based on their basic lexical components. It is used in computer programming and artificial intelligence, as well as natural language processing and interpretation. In more complex cases, lemmatization allows the computer to group words that share an inflected meaning but do not share a stem, such as "good" with phrases like "better" and "best."

### C. Data Visualization

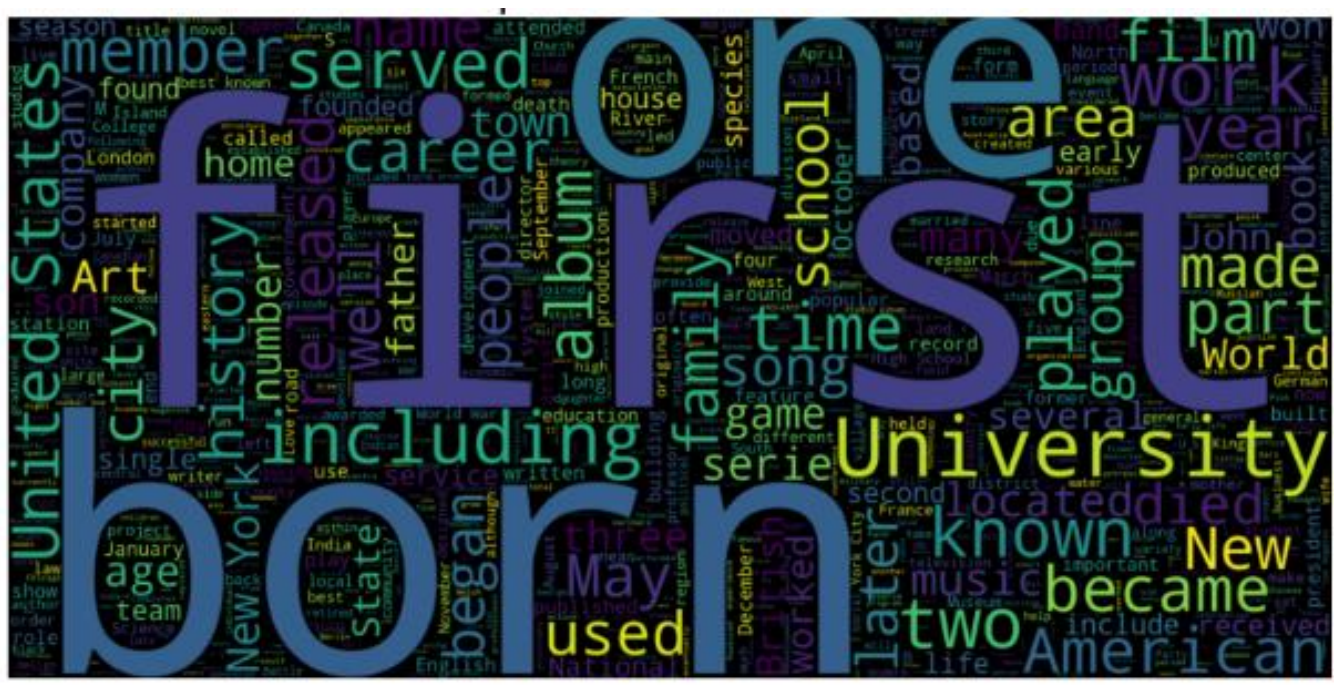

Fig. 2: Word cloud of text dataset

In figure 2, are showing word cloud of text in the dataset. Word clouds provide text data in different sizes depending on frequency or relevance. It helps you understand a text's main ideas quickly and easily. A word cloud, unlike a list or paragraph, enlarges words proportionately to their frequency, making the most important keywords stand out. This visualization method simplifies enormous volumes of textual data into an attractive manner, making it easy to see patterns. Data analysis, market research, and text summarization employ word clouds to identify key themes and patterns in a corpus of text.

### D. Feature Engineering

In natural language processing (NLP), CountVectorizer is used to extract token counts from text texts. By converting raw text input to numbers, machine learning algorithms can process it. Tokenize text documents, break them into words or terms, and count their occurrences using CountVectorizer. Thus, a sparse matrix with rows representing documents and columns representing unique terms in the corpus of documents is produced. Values in matrix cells represent phrase frequency in the document. The numerical representation of text data allows NLP tasks including text categorization, grouping, and information retrieval. Text processing pipelines employ CountVectorizer and other methods like term frequency-inverse document frequency (TF-IDF) to enhance NLP models.

### E. Model Generate

*BERT:*
Bidirectional Encoder Representations from Transformers, or BERT for short, is a cutting-edge language model that combines self-attention mechanisms with the Transformer architecture for comprehensive text processing. This algorithm operates with an exceptional 93% accuracy in our development. BERT is a cutting-edge model in natural language processing that uses pre-training tasks like Next Sentence Prediction and Masked Language Modelling, bidirectional training, and task-specific fine-tuning to obtain outstanding results. It excels at handling distant dependencies, collecting context-aware embeddings, and adjusting to different NLP applications.

*XGB:*
XGBClassifier, a powerful machine learning model, performs well in your assignment with 84.33% accuracy. The classifier's accuracy and recall statistics reveal its ability to distinguish human-written and AI-written material. Class 0 accuracy is 86%, indicating 86% of predicted instances are negative. The classification system detected 82% of class 0 instances, its recall. Class 1 also has 83% recall and 87% accuracy. The F1 ratings for the two categories reveal a decent accuracy-recall trade-off, making the machine learning approach robust. The confusion matrix reveals 246 real negatives, 260 genuine positives, 54 erroneous positives, and 40 erroneous negatives, illustrating the classifier's anticipated accuracy. Due to its high accuracy and precision-recall efficiency, the XGBClassifier is ideal for this project's classification task.

*SVM:*
This project's Support Vector Machine (SVM) classifier distinguishes AI-generated text from human-written material with 80.67% success. Accuracy and recall measures reveal the classifier's discriminating capacity. Class 0 accuracy is 79%, indicating 79% of anticipated instances are negative. Class 0 accuracy is 83%, meaning the classifier found 83% of true class 0 instances. Class 1 has 78% recall and 82% accuracy. Recall and accuracy are balanced in both groups' F1-scores. The disorientation matrix exhibits 249 genuine negatives, 235 real positives, 51 false positives, and 65 incorrect negatives to illustrate the classifier's prediction accuracy. The SVM classifier categorizes human-authored and AI-generated textual material with a median accuracy of 80.67%.

## IV. Results and Discussion

TABLE I. PERFORMANCES OF DIFFERENT CLASSIFIERS

| **Algorithm Name** | **Class** | **Precision** | **Recall** | **F1 Score** | **Accuracy** |
|---|---|---|---|---|---|
| XGB Classifier | 0 | 0.86 | 0.82 | 0.84 | 0.84 |
| | 1 | 0.83 | 0.87 | 0.85 | |
| SVM | 0 | 0.79 | 0.83 | 0.81 | 0.81 |
| | 1 | 0.82 | 0.78 | 0.80 | |

The evaluation of three distinct machine learning algorithms' effectiveness in identifying text created by artificial intelligence (AI): BERT, XGBoost (XGB), and Support Vector Machines (SVM). With an accuracy of 93%, BERT was the most accurate of these algorithms; XGBoost and SVM came in at 84% and 81%, respectively Google created BERT, or Bidirectional Encoder Representations from Transformers, a cutting-edge natural language processing (NLP) approach. It performs better in a variety of NLP tasks, such as text classification and sentiment analysis, by comprehending the context of words in a phrase by considering both the words that come before and after. To evaluate how well these algorithms, differentiate between material created by AI and content created by humans, it is essential to compare how well they recognize text written by AI. BERT's much greater accuracy indicates that it is better at identifying the subtleties and subtle patterns in text data that are suggestive of artificial intelligence (AI) development. On the other hand, while XGBoost and SVM exhibit reasonable accuracy, they may not be able to fully grasp the subtleties and complexity of language created by artificial intelligence to the same degree as BERT.

## V. Conclusion

Recent breakthroughs in natural language processing (NLP) may allow AI models to write like humans. This might have major ethical, legal, and societal consequences. An accurate AI detector model that can distinguish electronically generated text from human-written text is proposed in this paper. XGB Classifier, SVM, and BERT architecture deep learning models are used in our methodology. Our findings also reveal that the BERT identifies AI-generated information from human-provided information better than earlier models. Assess relevant research and analyze AI-generated text identification's present stage. Our testing showed that our technique works, with the BERT being the most likely response. We examine the research's social impacts, emphasizing industry benefits while addressing ethical and environmental sustainability challenges. This article's XGB classifier and SVM have 0.84 and 0.81 accuracy. The BERT model has the highest accuracy in this investigation at 0.93%.The examination of BERT, XGBoost (XGB), and Support Vector Machines' ability to detect AI-generated text. BERT was the most accurate at 93%, followed by XGBoost at 84% and SVM at 81%. Google developed cutting-edge NLP method BERT, or Bidirectional Encoder Representations from Transformers. It improves NLP tasks like text classification and sentiment analysis by understanding the context of words in a phrase by considering both preceding and following words. It is crucial to assess how effectively these algorithms detect AI-written language to determine how well they distinguish between AI and human-created information. BERT's higher accuracy suggests it can better recognize text data peculiarities that signify AI progress. XGBoost and SVM are accurate, however they may not understand the intricacies and complexity of artificial intelligence-generated language like BERT.